\documentclass[11pt, logo, copyright, numbering]{googledeepmind}

\pdftrailerid{redacted}

\makeatletter
\renewcommand\bibentry[1]{\nocite{#1}{\frenchspacing\@nameuse{BR@r@#1\@extra@b@citeb}}}
\makeatother

\usepackage{kantlipsum, lipsum}
\usepackage{dsfont}
\usepackage{gdm-colors}

\usepackage{microtype}
\usepackage{graphicx}
\usepackage{booktabs} 

\usepackage{subcaption}
\usepackage{caption}

\usepackage{hyperref}
\usepackage{algorithm}
\usepackage{algorithmic}

\usepackage[capitalize,noabbrev]{cleveref}

\usepackage{xspace}
\ifx\theorem\undefined

\theoremstyle{plain}
\newtheorem{theorem}{Theorem}[section]

\theoremstyle{definition}

\theoremstyle{remark}

\fi

\usepackage[textsize=tiny]{todonotes}
\usepackage{Definitions}

\newcommand{\algabb}{\textsc{MINC}\xspace}
\newcommand{\algname}{\text{Mutual Information Non-Contrastive}\xspace}

\usepackage[authoryear, sort&compress, round]{natbib}

\graphicspath{{figures/}}

\title{Representation Learning via Non-Contrastive Mutual Information}

\correspondingauthor{danielguo@google.com, bodai@google.com}

\reportnumber{} 

\renewcommand{\today}{2025-02-21}

\author[1]{Zhaohan Daniel Guo}
\author[1]{Bernardo Avila Pires}
\author[1]{Khimya Khetarpal}
\author[1]{Dale Schuurmans}
\author[1]{Bo Dai}

\affil[1]{Google DeepMind}

\begin{abstract}
    Labeling data is often very time consuming and expensive, leaving us with a majority of unlabeled data. Self-supervised representation learning methods such as SimCLR~\citep{chen2020simple} or BYOL~\citep{grill2020bootstrap} have been very successful at learning meaningful latent representations from unlabeled image data, resulting in much more general and transferable representations for downstream tasks. Broadly, self-supervised methods fall into two types: 1) Contrastive methods, such as SimCLR; and 2) Non-Contrastive methods, such as BYOL. Contrastive methods are generally trying to maximize mutual information between related data points, so they need to compare every data point to every other data point, resulting in high variance, and thus requiring large batch sizes to work well. Non-contrastive methods like BYOL have much lower variance as they do not need to make pairwise comparisons, but are much trickier to implement as they have the possibility of collapsing to a constant vector. In this paper, we aim to develop a self-supervised objective that combines the strength of both types. We start with a particular contrastive method called the Spectral Contrastive Loss~\citep{haochen2021provable,lu2024f}, and we convert it into a more general non-contrastive form; this removes the pairwise comparisons resulting in lower variance, but keeps the mutual information formulation of the contrastive method preventing collapse. We call our new objective the \algname~(\algabb) loss. We test~\algabb by learning image representations on ImageNet (similar to SimCLR and BYOL) and show that it consistently improves upon the Spectral Contrastive loss baseline.
\end{abstract}

\begin{document}

\maketitle

\section{Introduction}

In practice, it is often the case that we have a large dataset of unlabeled data and a smaller dataset of labeled data. This is because labeling data can be expensive due to time or cost, especially if we want high quality labels. In order to still take advantage of the large unlabeled dataset, self-supervised algorithms have been developed that first learn some kind of latent representation from the unlabeled data. The learned latent representation is then further used with the labeled data, either by training new layers on top, or fine-tuning. SimCLR~\citep{chen2020simple} and BYOL~\citep{grill2020bootstrap} are examples of this approach for large image datasets, and they have been shown to learn great representations in ImageNet. In addition to not needing labels, another advantage of self-supervised methods is that they learn more general representations. Because they are trained using unlabeled data, they learn representations that can better capture intrinsic qualities of the data, which are then much more transferable~\citep{chen2020simple}; then with fine-tuning or just a couple of new layers, they can quickly adapt to a desired downstream task.

Most of the effective self-supervised algorithms can be broadly split into two categories: 1) contrastive; and 2) non-contrastive. Contrastive losses have the following general idea: Pull together the embeddings of similar images closer, and push apart the embeddings of images that are different. A strength of many of these losses, such as InfoNCE~\citep{oord2018representation}, SimCLR, Spectral Contrastive~\citep{haochen2021provable}, $f$-MICL~\citep{lu2024f}, is that they are theoretically based in maximizing the mutual information between related data points. This gives a solid theoretical foundation for analyzing what kind of representation is learned; for example, the Spectral Contrastive loss learns a type of spectral decomposition of the data. However these contrastive losses have an inherent weakness; because they are concerned about comparing data points to one another, they require every data point to eventually be compared with every other data point, resulting in a quadratic dependence on the size of the dataset. This quadratic dependence means that the objective function has very high variance, which means large batch sizes are required to minimize these losses effectively.

Non-contrastive losses like BYOL and SimSiam do not have the quadratic dependence because they do not need to compare all data points to each other. Furthermore, they currently achieve state-of-the-art (when properly tuned), outperforming many contrastive algorithms. They only have a linear dependence, meaning much lower variance. However the tradeoff is that, theoretically and empirically, they are more brittle and prone to collapse, \ie, a constant representation minimizes their losses. They require specific empirical tweaks such as additional layers of MLP, target networks, and are not great to use with weight decay for regularization, as it can increase the chance of collapse or partial collapse. Recent theoretical understanding has shown when they do not collapse but learn a meaningful representation, however the theoretical assumptions do not hold in practice, and so the issue of collapse still remains.

In this paper, we propose a new non-contrastive objective, the \algname (\algabb) loss, which retains the non-collapse guarantees enjoyed by contrastive losses, while removing the quadratic data dependence of contrastive methods.
Instead of comparing every data point to every other data point, the \algabb loss compares each data point to a summary of all other data points. We derive \algabb starting with the Spectral Contrastive Loss, and reformulating it using the lens of power iteration in combination with insights from the Generalized Hebbian Algorithm~\citep{sanger1989optimal}. The power iteration formulation also acts as a bridge to connect \algabb with BYOL using a linear predictor.

To evaluate \algabb, we follow the same setup as for SimCLR and BYOL and test the representation learning in ImageNet. We first learn a representation from the unlabeled dataset, and fix it. Then we train a linear classifier on top of the fixed representation to evaluate how effective the learned representation is for image classification. Our experiments show that \algabb is consistently improving upon the Spectral Contrastive Loss baseline, and matches up to BYOL with a linear predictor. While it does not quite match up to the regular BYOL that uses a non-linear predictor, these results do show the promising direction of turning a contrastive loss into a non-contrastive loss to get the best of both worlds. Perhaps a better contrastive loss is needed as a foundation to connect to the non-linear BYOL.

\section{Preliminaries}
We first introduce the generalization of InfoNCE-based contrastive learning~\citep{lu2024f} through $f$-divergences, which includes the SimCLR~\citep{chen2020simple}, InfoNCE~\citep{oord2018representation} and Spectral Contrastive loss~\citep{haochen2021provable} as special cases. 
We further provide a brief introduction to BYOL~\citep{grill2020bootstrap}, the representative non-contrastive representation learning method, which not only inspired our algorithm, but also whose connection to mutual information will be revealed during our algorithm derivation. 

\noindent{\bf $f$-InfoNCE.} Consider a pair of random variables $(X, X')$ both taking values in $\mathcal{X}$ with joint density $p(x,x')$ and marginal densities $p(x)$ and $p(x')$, respectively. The $f$-mutual information between $X$ and $X'$ is defined as 
\begin{equation}\label{eq:f_mi}
    I_f\rbr{X, X'}\defeq \int f\rbr{\frac{p(x, x')}{p(x)p(x')}}p(x)p(x')dx dx',
\end{equation}
where $f: \RR_+\rightarrow \RR$ is a convex function with $f(1) = 0$. In fact, the $f$-mutual information~(MI) is the $f$-divergence between $p(x, x')$ and $p(x)p(x')$, which intuitively characterizes the entangle part between $X$ and $X'$ only in joint distribution. Letting $f^* : \RR \rightarrow \RR$ be the Fenchel duality (convex conjugate) of $f$~\citep{nguyen2010estimating,nachum2020reinforcement}, we have
\[
    f(u) = \max_{t\in \RR} \rbr{u\cdot t - f^*(t)},
\]
which reformulates the $f$-MI as
\begin{equation}\label{eq:f_dual}
\begin{aligned}
    I_f\rbr{X, X'} = \max_{t : \mathcal{X} \times \mathcal{X} \rightarrow \RR}\,\, 
    \EE_{p(x, x')}\sbr{t(x, x')} - \EE_{p(x)p(x')}\sbr{f^*\rbr{t(x, x')}}.
\end{aligned}
\end{equation}
Under this formulation, the representation learning consists of parametrizing $t$ on the right-hand side using of an embedding function $\phi$, and finding a $\phi$ that maximizes the expression.
Since we experiment with image representations, for ease of explanation, we will assume that the data $x$ are pixel images, and that $\phi$ is a ResNet neural network that encodes the image into a latent vector. Plugging the parametrization of $t\rbr{x, x'}$ with representation of $\phi\rbr{x}$ and $\phi\rbr{x'}$ into~\eqref{eq:f_dual}, with different choices of $f$
but also recover several existing contrastive representation learning losses~\citep{lu2024f}, therefore, revealing the mutual information essence in representation learning. 

Specifically, with $f\rbr{u} = (u-1)^2$ and the corresponding $f^*\rbr{t} = \frac{t^2}{4} + t$ in~\eqref{eq:f_dual}, we consider $t(x, x') = 2(\phi\rbr{x}^\top \phi\rbr{x'} - 1)$, leading to the representation learning objective (ignoring constants)
\begin{equation}\label{eq:spectral_contrastive}
\begin{aligned}
    \max_{\phi}\,\, 
    2\EE_{p\rbr{x, x'}}\sbr{\phi(x)^\top \phi(x')}  - \EE_{p(x)p(x')}\sbr{\rbr{\phi(x)^\top \phi(x')}^2},
\end{aligned}
\end{equation}
which is the Spectral Contrastive loss introduced by~\citet{haochen2021provable}. This choice of $f$ corresponds to the commonly known $\chi^2$-divergence. Similarly, SimCLR~\citep{chen2020simple} and InfoNCE~\citep{oord2018representation} can be related to this framework using a variational form of~\eqref{eq:f_dual} with mutual information from KL-divergence through $f(u) = u\log u $, as discussed by~\citet{lu2024f}.

\noindent{\bf BYOL.} A drawback of contrastive losses for representation learning is that they require careful treatment of how to compare every sample to every other sample~\citep{wu2017sampling}, and large batch sizes~\citep{chen2020simple,tian2020makes}.
Non-contrastive representation learning methods like BYOL~\citep{grill2020bootstrap} aim to circumvent these issues. Specifically, BYOL is proposed to learn the representation $\phi$ through direct prediction of self-generated targets:
\begin{equation}\label{eq:byol}
    \min_{\phi, \Lambda}\,\, \EE_{p(x, x')}\sbr{\nbr{\Lambda\rbr{\phi\rbr{x}} - \text{sg}\rbr{\phi\rbr{x'}}}^2}, 
\end{equation}
where $\Lambda$ is the predictor, and $\mathrm{sg}\rbr{\phi\rbr{x'}}$ are the prediction targets (with a stop-gradient) used for representation learning. Often in practice, the prediction targets also come from a slow moving target network. Although BYOL achieves state-of-the-art performance, the objective has a quirk that a trivial $\phi(\cdot) = 0$, often referred to as a collapsed representation, is optimal.
In practice, BYOL has been shown to avoids collapse and learn meaningful representations, and there have been several attempts to investigate why.
\citet{tang2022understanding} characterized the optimal solution of the BYOL learning dynamics with semi-gradient optimization, while \citet{richemond2023edge} revealed the orthonormalization effects in the updates through spectral view and Riemannian gradients. These theoretical approaches showed that BYOL is related to some kind of spectral decomposition of the data. However, these analyses make many assumptions that are not present in practice, in particular they both analyze the case of a linear predictor $\Lambda$ as opposed to an MLP predictor used in practice. Thus, while BYOL does not have issues related to negative examples, in practice it still requires careful execution to prevent collapse.

\section{Mutual Information through Non-Contrastive Loss}

As discussed, contrastive losses like the Spectral Contrastive loss are founded in the $f$-mutual information framework and so have a solid theoretical basis. 
They are not prone to representation collapse, but are held back by the quadratic dependence on the number of samples.
In this section, we introduce the \emph{\algname~(\algabb)} loss, which retains the non-collapse guarantees enjoyed by $f$-MI contrastive losses, while, at the same time, avoiding the quadratic dependence by being \emph{non-contrastive}.

\subsection{First Attempt at Non-Contrastive}

A simple first attempt to convert the Spectral Contrastive loss~\eqref{eq:spectral_contrastive} to a non-contrastive loss starts with rearranging:
\begin{align*}
    \max_{\phi} \,\, 2\EE_{p\rbr{x, x'}}\sbr{\phi\rbr{x}^\top \phi\rbr{x'}} - \EE_{p(x')}\sbr{\phi\rbr{x'}^\top \EE_{p(x)} \sbr{\phi\rbr{x} \phi^\top\rbr{x}} \phi\rbr{x'} },
\end{align*}
which has separated the quadratic dependence on the dataset (in the second term of~\eqref{eq:spectral_contrastive}) into an inner and outer expectation. Then we can introduce an auxiliary variable with a constraint to completely decouple the inner expectation:
\begin{align}
    \max_{\phi} &\,\, 2\EE_{p\rbr{x, x'}}\sbr{\phi\rbr{x}^\top \phi\rbr{x'}} - \EE_{p(x')}\sbr{\phi\rbr{x'}^\top \Lambda \phi\rbr{x'} } \nonumber\\
    &\text{subject to } \quad \Lambda = \EE_{p(x)} \sbr{\phi\rbr{x} \phi^\top\rbr{x}}, \label{eq:firstaux}
\end{align}
and this successfully becomes a non-contrastive objective, because it removes the quadratic dependence (the double-integral weighted by $p(x)p(x')$) through the auxiliary matrix $\Lambda$. Essentially, $\Lambda$ has become a \emph{summary} of the statistics of $\phi(x)$, allowing $\phi(x')$ to be compared with this summary matrix instead of every other point.

To tackle this new constrained objective, we will first re-contextualize the Spectral Contrastive loss in an eigen-decomposition view. Then we will employ the simple and efficient power iteration method for eigen-decomposition to help derive a new non-contrastive objective.

\subsection{Spectral Contrastive as Eigen-Decomposition}

Here we re-contextualize the Spectral Contrastive loss~\eqref{eq:spectral_contrastive} through the eigen-decomposition perspective originally adopted by~\citet{haochen2021provable, ren2022spectral, ren2024spectral}. Given embeddings $\phi(x), \phi(x')$, and assuming a finite number of data point pairs $N$, let $F: \RR^{N} \times \RR^d$ be the matrix where we stack all the scaled $\sqrt{p(x)}\phi(x)$ as rows (and similarly for $F'$ and $x'$). Then we can formulate the following low-rank matrix decomposition problem:
\begin{align}\label{eq:matrix-factorization}
    \min_{\phi} \Vert M - FF'^\top \Vert_\mathrm{F}^2,
\end{align}
where $M_{ij} \defeq \frac{p(x_i, x_j')}{\sqrt{p(x_i)p(x_j')}}$, is a quantity related to the mutual information of the dataset. The solution is then for $F$ to share the eigen-space of $M$ (more precisely, $F$ will be the matrix of eigenvectors multiplied by the diagonal matrix of the square root of the eigenvalues). This means the solution satisfies (in pointwise notation):
\begin{align}\label{eq:factorization}
    \frac{p(x, x')}{\sqrt{p(x)p(x')}} & = \rbr{\sqrt{p(x)}\phi\rbr{x}^\top} \rbr{\sqrt{p(x')} \phi\rbr{x'}},
\end{align}
Then in the more general case of~\eqref{eq:matrix-factorization} we have:
{\small
\begin{align*}
    \min_\phi&\int\rbr{ \frac{p(x, x')}{\sqrt{p(x)p(x')}} -  \sqrt{p(x)p(x')}\phi\rbr{x}^\top \phi\rbr{x'}}^2 dx dx' \\
    =&\underbrace{\int\frac{p^2\rbr{x',x}}{p\rbr{x'}p\rbr{x}} dx'dx}_{\texttt{constant}} - 2\EE_{p\rbr{x, x'}}\sbr{\phi\rbr{x}^\top \phi\rbr{x'}}   + \EE_{p(x)p(x')}\sbr{\rbr{\phi\rbr{x}^\top \phi\rbr{x'}}^2}, 
\end{align*}}
giving us back the Spectral Contrastive loss.

\subsection{From Contrastive to Non-Contrastive Through Power Iteration }

Given the eigen-decomposition view of the Spectral Contrastive loss in~\eqref{eq:factorization}, it is very natural to exploit the rich numerical linear algebra literature for faster eigen-decomposition algorithms, among which power iteration~\citep{golub2013matrix,trefethen2022numerical} is a simple yet effective algorithm for eigen-deomposition. It has been proved the convergence rate of subspace iteration is exponential, therefore motivating us to extend power iteration for representation learning. 

Recall the properties of eigen-decomposition, first we have an orthogonality condition, which we can write as 
\begin{equation}\label{eq:orth_cond}
    \int \phi\rbr{x}\phi\rbr{x}^\top p(x)dx =\int \phi\rbr{x'}\phi\rbr{x'}^\top p(x')dx' = \Lambda,
\end{equation}
where $\Lambda = \diag\rbr{[\lambda_i]_{i=1}^d}$ as the diagonal matrix constructed by the eigenvalues. Note that we are reusing the symbol $\Lambda$ that was for the auxiliary matrix in~\eqref{eq:firstaux} because they will end up matching. Meanwhile, the eigen-function satisfies the following fixed point equation for all $x'$:
\begin{align}\label{eq:eigen_fac}
    \int \frac{p(x, x')}{\sqrt{p(x)p(x')}} \sqrt{p(x)}\phi\rbr{x}dx = \Lambda \sqrt{p(x')}\phi\rbr{x'}. 
\end{align}
Power iteration works by iterating a fixed point update through~\eqref{eq:eigen_fac} while ensuring the orthogonality condition~\eqref{eq:orth_cond}. Concretely we have the following update rules for each iteration $t$:
\begin{subequations}\label{eq:conceptual_pi}
\begin{align}
        \Lambda_{t+1} &= \EE_{p(x)}\sbr{\phi_t\rbr{x}\phi_t\rbr{x}^\top},\label{eq:lambda} \\
        \Lambda_{t+1}\sqrt{p(x')}\psi_{t+1}\rbr{x'} &=  \int \frac{p(x, x')}{\sqrt{p(x')}} \phi_t\rbr{x}dx, \label{eq:var_iter}\\
        \sqrt{p(x')}\phi_{t+1}\rbr{x'} &=\hspace{-1mm} \textbf{orth}\rbr{\sqrt{p(x')}\psi_{t+1}\rbr{x'}}. \label{eq:orth}
\end{align}
\end{subequations}
Note how~\eqref{eq:lambda} is actually the same constraint we introduce in~\eqref{eq:firstaux}; this means power iteration can give us an explicit and efficient approach that subsumes the constrained non-contrastive objective.

The exact integrals in the first two updates and the orthogonalization in the third update are intractable, however we will approximate them effectively to arrive at a practical power iteration procedure.

\paragraph{Moving Average for~\eqref{eq:lambda}.} Given  $\phi_t$, $\Lambda_{t+1}$ is formulated as an expectation, therefore, it can be simply approximated by Monte Carlo estimation, \ie, 
\begin{equation*}
    \textstyle
    \Lambda_{t+1}\approx \frac{1}{n}\sum_{i=1}^n \phi_t(x_i)\phi_t(x_i)^\top,\quad \cbr{x_i}_{i=1}^n\sim p\rbr{\cdot}. 
\end{equation*}
However, this Monte Carlo ignores our iterative estimation procedure. We convert the estimation to an optimization, which leads to a moving average procedure, \ie,
\begin{equation}
    \textstyle
    \Lambda_{t+1} = \argmin_{\Lambda}\,\,\EE_{p(x)}\sbr{\nbr{\Lambda - \phi_t(x)\phi_t(x)^\top}_2^2}
    + \eta \nbr{\Lambda - \Lambda_t}_\mathrm{F}^2,
\end{equation}
inducing the exponential moving average (EMA) update
\begin{equation}\label{eq:lambda_update}
    \Lambda_{t+1} = \beta\Lambda_t + (1 - \beta) \EE_{p(x)}\sbr{\phi_t(x)\phi_t(x)^\top}
\end{equation}
with $\beta = \frac{\eta}{1 + \eta}$. 
    
\paragraph{Orthogonal Variational Iteration via GHA for~\eqref{eq:var_iter} and~\eqref{eq:orth}.} 
With $\Lambda_{t+1}$ being updated by an EMA, we can recast the iteration update for $\psi$ as an optimization by matching LHS and RHS of~\eqref{eq:conceptual_pi} via Mahalanobis distance with $\Lambda_{t+1}^{-1}$, \ie, 
\begin{align*}
    \textstyle
    \hspace{-3mm}&\min_{\psi}\,\, \int \sbr{\nbr{\int \frac{p(x, x')}{\sqrt{p(x')}}\phi_t\rbr{x}dx  - \Lambda_{t+1}\sqrt{p(x')}\psi\rbr{x'}}_{\Lambda_{t+1}^{-1}}^2}dx'\nonumber\\
    \propto& \underbrace{- 2\EE_{p(x, x')}\sbr{\phi_t\rbr{x}^\top\psi\rbr{x'}} + \EE_{p(x')}\sbr{\psi\rbr{x'}^\top \Lambda_{t+1} \psi\rbr{x'}}}_{\ell\rbr{\psi}}.
\end{align*}
where its corresponding gradient update is
\begin{equation}\label{eq:psi_grad}
    \frac{1}{2}\nabla_\psi \ell\rbr{\psi}= - \EE_{p(x, x')}\sbr{\phi_t\rbr{x}^\top\nabla_\psi\psi\rbr{x'}} + \EE_{p(x')}\sbr{\psi\rbr{x'}^\top { \Lambda_{t+1}} \nabla_\psi\psi\rbr{x'}},
\end{equation}
to approximate the update rule for $\psi$ related to \eqref{eq:var_iter}. 

To obtain $\phi_{t+1}$ via \eqref{eq:orth} we need to orthogonalize $\psi_{t+1}$. One natural choice is Gram-Schmidt process, but is computationally expensive (or intractable).
We consider to relax the strict orthogonal condition, only maintaining \emph{asymptotic orthogonality}. 
This can be achieved through the generalized Hebbian algorithm~\citep[GHA;][]{sanger1989optimal,xie2015scale}, which ensures the orthogonality in an asymptotic sense.
By combining the GHA update rule and the gradient update of $\psi$ from~\eqref{eq:psi_grad}, 
we can have $\psi_{t+1}$ asymptotically orthogonal by uptading it with suitably chosen step size $\kappa_t$ as
\begin{equation*}
    \psi_{t+1} = \psi_{t} + \kappa_t\bigg( \EE_{p(x, x')}\sbr{\phi_t\rbr{x}^\top\nabla_\psi\psi_t\rbr{x'}} - \EE_{p(x')}\sbr{\psi_t\rbr{x'}^\top {\color{blue}\texttt{LT}}\sbr{ \Lambda_{t+1}} \nabla_\psi\psi_t\rbr{x'}} \bigg),
\end{equation*}
where $\texttt{LT}\sbr{\cdot}$ stands for making matrix lower triangular by setting all elements above the diagonal of
its matrix argument to zero.

Because the GHA rule ensures asymptotic orthogonality, we skip the explicit orthogonalization in~\eqref{eq:orth}, which gives $\phi_{t+1} = \psi_{t+1}$.
Therefore, we can fuse~\eqref{eq:psi_grad} and~\eqref{eq:orth} into a direct update for $\phi$:
\begin{equation}\label{eq:phi_update}
    \phi_{t+1} = \phi_{t} + \kappa_t\bigg( \EE_{p(x, x')}\sbr{\phi_t\rbr{x}^\top\nabla_\phi\phi_t\rbr{x'}} - \EE_{p(x')}\sbr{\phi_t\rbr{x'}^\top {\color{blue}\texttt{LT}}\sbr{ \Lambda_{t+1}} \nabla_\phi\phi_t\rbr{x'}}\bigg).
\end{equation}

The combined updates for $\Lambda$ and $\phi$ from~\eqref{eq:lambda_update} and~\eqref{eq:phi_update}, respectively, give us a tractable, non-contrastive power iteration method.
This method is also an asymptotic non-contrastive approximation of gradient descent on the Spectral Contrastive loss. 
To see this, note that the gradient of the Spectral Contrastive loss~\eqref{eq:spectral_contrastive} with respect to $\phi$ is, 
\begin{equation*}
    - 2\bigg( \EE_{p(x, x')}\sbr{\phi_t\rbr{x}^\top\nabla_\phi\phi\rbr{x'}} + \EE_{p(x')p(x)}\sbr{\phi\rbr{x'}^\top{\color{blue}\rbr{\phi\rbr{x}\phi\rbr{x}^\top}}  \nabla_\phi\phi\rbr{x'}} \bigg).
\end{equation*}
Comparing this with the update rules in~\eqref{eq:lambda_update} and~\eqref{eq:phi_update}, we can recover the gradient of the Spectral Contrastive loss with $\EE_{p(x)} \sbr{\phi_t\rbr{x} \phi_t^\top\rbr{x}}$ approximated by $\texttt{LT}\sbr{ \Lambda_{t+1}}$.

This asymptotic non-contrastive approximation not only holds for the $\chi^2$-MI discussed so far, but it can also be extended to other $f$-MI objectives, as we show next.

\subsection{Generalization for $f$-MI Representation}

The general $f$-MI objective is restated as follows with an embedding network $\phi$, as well as a choice of a scalar encoding function $t : \RR_+ \rightarrow \RR$, \ie,
\begin{align}
    I_f\rbr{X, X'} &= \max_{\phi}\,\, \EE_{p(x, x')}\sbr{t \rbr{\phi(x)^\top \phi(x')}}  - \EE_{p(x)p(x')}\sbr{f^*\rbr{t\rbr{\phi(x)^\top \phi(x')}}}.  
\end{align}
The first term of $\EE_{p(x, x')}\sbr{t \rbr{\phi(x)^\top \phi(x')}}$ can be put aside, because it does not have the quadratic variance issue. So we focus on the second term $\EE_{p(x)p(x')}\sbr{f^*\rbr{t\rbr{\phi(x)^\top \phi(x')}}}$. For the $\chi^2$-divergence, we already know that we get a squared dot product term for this second term, which allows us to make the non-contrastive conversion. In the more general $f$-divergence case, it suffices then to pick $f^*$ and $t$ such that we get back to a squared dot product form again.

We consider a family of $f$-divergences, the $\alpha$-divergences. The $\alpha$-divergences generalize many common divergences through a real parameter $\alpha$. For example, when $\alpha \in [0, 1, 2]$ we get the reverse KL, KL, and (Pearson) $\chi^2$ divergences respectively. This means the Spectral Contrastive loss corresponds to $\alpha=2$. We know $f$ and $f^*$ and we can pick $t$ as follows:
\begin{align}
f_{\alpha}(y) &= \frac{\left( y^\alpha - 1 \right) - \alpha(y - 1)}{\alpha \left( \alpha - 1 \right)} \\
f^\star_\alpha(t) &= \frac{1}{\alpha} \left| 1 + \left( \alpha - 1 \right)t\right|^{\frac{\alpha}{\alpha - 1}} - \frac{1}{\alpha} \\
    t_{\alpha}(u) &= \operatorname{sign}(u)\frac{1}{\alpha - 1}\left\vert \sqrt{\frac{\alpha}{2}} u \right\vert^\frac{2(\alpha-1)}{\alpha}  - \frac{1}{\alpha - 1}. \label{eq:t_alpha}
\end{align}
Plugging these in leads to the following converted objective:
\begin{align}
    \max_{\phi} \,\, \EE_{p(x, x')}\sbr{t_{\alpha} \rbr{\phi(x)^\top \phi(x')}}  - \frac{1}{2}\EE_{p(x)p(x')}\sbr{\rbr{\phi(x)^\top \phi(x')}^2} + \frac{1}{\alpha}.
\end{align}
Now the second term has the square dot product form as needed, which we can rewrite with the auxiliary matrix $\Lambda$, giving us the corresponding generalized objective (safely ignoring the constant $1/\alpha$):
\begin{align}
\label{eq:minc_pre}
    \max_{\phi} \,\, \EE_{p(x, x')}\sbr{t_{\alpha} \rbr{\phi(x)^\top \phi(x')}} - \frac{1}{2}\EE_{p(x')}\sbr{\rbr{\phi(x')^\top \textcolor{blue}{\texttt{LT} \sbr{\Lambda}} \phi(x')}}.
\end{align}
where $\Lambda$ is updated through the EMA update rule~\eqref{eq:lambda_update}.

\subsection{The \algabb Objective}

There are two more additions that we make to the objective~\cref{eq:minc_pre} to further improve its empirical performance. The first is to normalize the embedding vector $\phi(x)$ so that it has Euclidean norm 1. This means the dot product is now a cosine similarity. Because this will constrain the magnitude of the dot product, it becomes important to allow the dot product to be scaled by a separate hyperparameter that we call the \emph{inner scale} $s$.
\begin{align}
    \max_{\phi} \,\, \EE_{p(x, x')}\sbr{t_{\alpha} \rbr{\textcolor{blue}{s} \phi(x)^\top \phi(x')}}  - \frac{1}{2}\EE_{p(x')}\sbr{\textcolor{blue}{s^2} \rbr{\phi(x')^\top \texttt{LT} \sbr{\Lambda} \phi(x')}}. \nonumber
\end{align}
This modification is also done in the Spectral Contrastive loss and SimCLR, though they use $1/s$ as the hyperparameter and denote it as the temperature.

The final addition we make is to use a target network $\phi_{\mathrm{target}}$ for $x$ and $\Lambda$, which gives the following objective and update rule ($t_\alpha$ is defined in equation~\eqref{eq:t_alpha}):
\begin{align}
\label{eq:minc_target}
    \max_{\phi} &\,\, \EE_{p(x, x')}\sbr{t_{\alpha} \rbr{s \phi_{\mathrm{target}}(x)^\top \phi(x')}}  - \frac{1}{2}\EE_{p(x')}\sbr{s^2 \rbr{\phi(x')^\top \texttt{LT} \sbr{\Lambda} \phi(x')}} \\
    \label{eq:lambda_target_update}
    \Lambda_{t+1} &= \beta\Lambda_t + (1 - \beta) \EE_{p(x)}\sbr{\phi_{\mathrm{target},t}(x)\phi_{\mathrm{target},t}(x)^\top}
\end{align}
This is similar to BYOL in that $\phi_{\mathrm{target}}$ is a slow, EMA updated version of $\phi$. We found this target network to both improve performance and increase learning stability under large learning rates. Putting it together we get~\cref{alg:minc}.

\begin{algorithm}[ht]
   \caption{\algabb}
   \label{alg:minc}
\begin{algorithmic}
   \STATE {\bfseries Input:} batch of data $\{(x, x')\}$, $\alpha$-divergence $\alpha$, inner scale $s$, auxiliary EMA $\beta$, neural network embedding $\phi$, target network EMA $\gamma$
   \STATE Initialize target network $\phi_{\mathrm{target},0} = \phi$
   \STATE Initialize auxiliary matrix $\Lambda_0 = 0$
   \FOR{$i=1$ {\bfseries to} $\dots$}
   \STATE Sample minibatch of data $B = \{(x_j, x_j')\}$
   \STATE Update $\Lambda_i$ through~\cref{eq:lambda_target_update} using $B$
   \STATE Update $\phi_i$ through gradient of~\cref{eq:minc_target} using $B$
   \STATE Update target network $\phi_{\mathrm{target},i}$ with EMA $\gamma$
   \ENDFOR
\end{algorithmic}
\end{algorithm}

\subsection{More Theoretical Connections}
\label{sec:moretheory}

\paragraph{Remark (Choice of Matching Metric):} In~\eqref{eq:phi_update}, we consider the Mahalanobis distance with $\Lambda_{t+1}^{-1}$. In fact, one can also select vanilla $L_2$ distance, \ie, 
{\allowdisplaybreaks
\begin{align*}
    & \int \sbr{\nbr{\int \frac{p(x, x')}{\sqrt{p(x')}}\phi_t\rbr{x}dx  - \Lambda_{t+1}\sqrt{p(x')}\phi\rbr{x'}}^2}dx'\nonumber\\
    \propto& - 2\EE_{p(x, x')}\sbr{\phi_t\rbr{x}^\top\Lambda_{t+1}\phi\rbr{x'}} + \EE_{p(x')}\sbr{\phi\rbr{x'}^\top \Lambda_{t+1}^\top\Lambda_{t+1} \phi\rbr{x'}}.
\end{align*}}
Comparing to~\eqref{eq:phi_update}, the $L_2$ distance leads to a similar objective, which is also stochastic gradient descent compatible, but with worse condition number. We can consider other metrics, which is out of the scope of the paper and we leave for future research.

\paragraph{Remark (Connection to BYOL and the variants):} 
BYOL~\citep{grill2020bootstrap} and its variants~\citep{richemond2023edge,tian2021understanding,wang2021towards} can be recast as variants in implementing power iteration. The major difference between \algabb and BYOL and its variants mainly lies in the update rule of $\Lambda$. We emphasize these update rules in BYOL~\citep{grill2020bootstrap} and its variants~\citep{richemond2023edge,tian2021understanding,wang2021towards} can also be derived by exploiting different properties of $\Lambda$. For example, as we show in~\eqref{eq:eigen_fac}, $\Lambda$ plays as the eigenvalue matrix in stationary solution, therefore, it should satisfy the variational characteristic of eigenvalues, \ie, 
\begin{align*}
    & \Lambda = \argmin_A \int \sbr{\nbr{\int \frac{p(x, x')}{\sqrt{p(x')}}\phi\rbr{x}dx  - A\sqrt{p(x')}\phi\rbr{x'}}^2}dx'\nonumber\\
    &\propto - 2\EE_{p(x, x')}\sbr{\phi_t\rbr{x}^\top A\phi\rbr{x'}} + \EE_{p(x')}\sbr{\phi\rbr{x'}^\top A^\top A \phi\rbr{x'}},
\end{align*}
which leads to the update rule in BYOL with a linear predictor. Similarly, we can also derive other update rule for $\Lambda$ through the properties of eigen-values, as shown in~\citet{richemond2023edge}. 

\section{Experiments}\label{sec:experiments}

We follow the same experimental procedures as in SimCLR~\citep{chen2020simple} and BYOL~\citep{grill2020bootstrap}. We train latent image representations in ImageNet using a standard ResNet-50 network~\citep{he2016deep} and our objective, with standard random image augmentations and without labels, as in SimCLR and BYOL. On top of the ResNet, we have a projector MLP akin to SimCLR, with output sizes $(2048, 2048, 2048)$ i.e.~two hidden layers with an output embedding size of $2048$.  The learned representations are evaluated by training a separate linear classifier on top of the frozen learned representations on a validation or test split. The representation that is evaluated is the immediate output of the ResNet, before the projector. For training we follow the same initial learning rate of $0.3$ with $10$ epochs of linear warmup, followed by cosine decay with the LARS optimizer and weight decay of $10^{-4}$, identical to SimCLR. When using a target network, we use an exponential moving average update with decay rate of $\gamma=0.996$, same as for BYOL, except we use a fixed decay of $0.996$ with no schedule. Unless otherwise specified, we use an auxiliary EMA $\beta=0.8$.  More details in the appendix. We train for $300$ epochs and $1$ seed due to computational cost.

\subsection{Main Result}

For our main results, we pick the $\alpha=2$ ($\chi^2$-divergence), as it was the best-performing and simplest compared to other nearby values of $\alpha$ (see the ablations in \cref{sec:ablations} and~\cref{fig:alphadivergence} for more details).

\begin{figure}[ht]
   \begin{center}
    \begin{subfigure}[t]{0.45\textwidth}
        \includegraphics[width=\columnwidth]{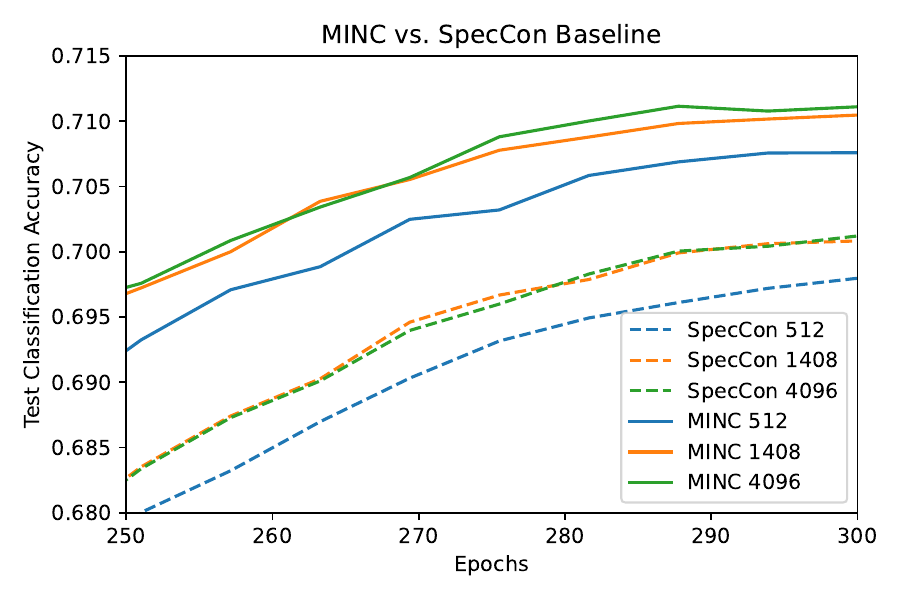}
    \caption{\algabb vs the SpecCon Baseline. \algabb improves consistently across all batch sizes.}
    \label{fig:main1}
    \end{subfigure}
    \hspace{4mm}
    \begin{subfigure}[t]{0.45\textwidth}
        \includegraphics[width=\columnwidth]{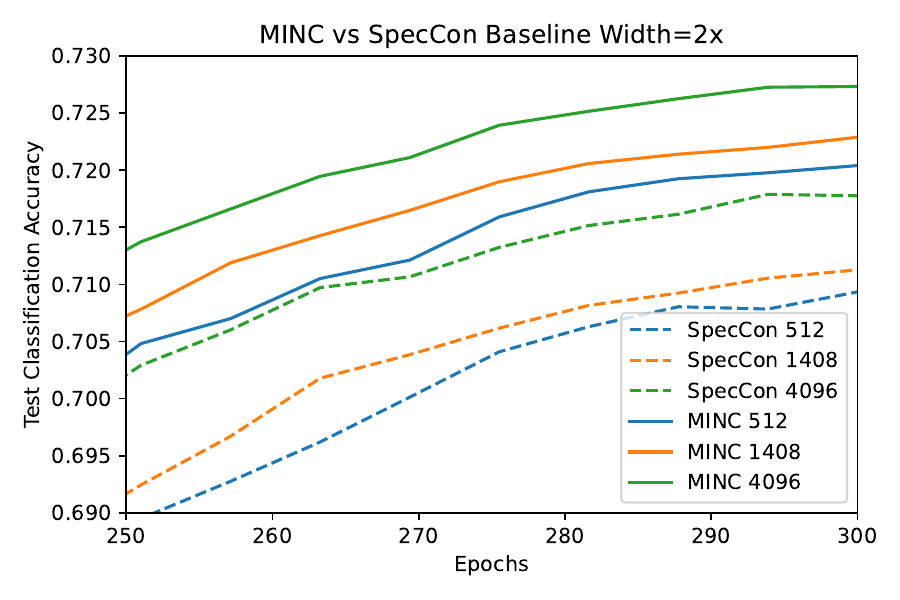}
    \caption{\algabb vs the SpecCon Baseline but doubling the width of the ResNet-50. \algabb still improves across the board.}
    \label{fig:main2}
    \end{subfigure}
    \end{center}
    \vskip -0.2in
\end{figure}

\begin{table}[ht]
\begin{center}
\caption{Main Result: Test Classification (Top-1) Accuracy}
\label{tbl:main}
\begin{small}
\begin{sc}
\begin{tabular}{ccc}
\toprule
Method & ResNet50 & ResNet50 2x \\
\midrule
SpecCon $512$ & $0.698$ & $0.709$ \\
\algabb $512$ & $0.708$ & $0.720$ \\
\midrule
SpecCon $1408$ & $0.701$ & $0.711$ \\
\algabb $1408$ & $0.710$ & $0.723$ \\
\midrule
SpecCon $4096$ & $0.701$ & $0.718$ \\
\algabb $4096$ & $0.711$ & $0.727$ \\
\midrule
SimCLR & $0.693$ & $0.742$ \\
MoCo v2 & $0.711$ & --- \\
SimSiam & $0.713$ & --- \\
Linear BYOL & $0.715$ & --- \\
Linear $\text{NS}^2$ BYOL & $0.722$ & --- \\
BYOL & $0.743$  & $0.774$ \\
\bottomrule
\end{tabular}
\end{sc}
\end{small}
\end{center}
\end{table}

Our main result is shown in~\cref{tbl:main}, with curves in~\cref{fig:main1,fig:main2}, where we compare our \algabb vs. a SpecCon baseline across three different batch sizes of $512$, $1408$ and $4096$. These results are evaluated using the ImageNet test split, and trained on both the train and validation splits. We also include reported results from SimCLR, MoCo v2~\citep{he2020momentum}, SimSiam, Linear BYOL (the predictor is linear and trained), Linear $\text{NS}^2$ BYOL~\citep{richemond2023edge} and BYOL~\citep{grill2020bootstrap}, for reference. \algabb consistently outperforms the Spectral Contrastive baseline across all batch sizes and for both network sizes (standard and $2\mathrm{x}$). This shows that \algabb is a robust improvement over the baseline Spectral Contrastive loss.

Compared with other methods, it outperforms SimCLR, another contrastive method, and almost matches some non-contrastive methods such as SimSiam and Linear BYOL. Performing similarly to Linear BYOL suggests that the power iteration framing and connection (\cref{sec:moretheory}) of \algabb to Linear BYOL may be very illuminating and useful for further development. However \algabb is still not able to catch up with the regular BYOL that uses a non-linear predictor, which shows that perhaps the Spectral Contrastive loss (and other contrastive losses) may still be limited in representational power. This leaves the door open for better contrastive methods to be used and subsequently converted to improved non-contrastive objectives.

\subsection{Ablations}
\label{sec:ablations}

\algabb is made up of three main components: 1) the \algabb objective with the auxiliary $\Lambda$; 2) The GHA update rule with the lower triangular transformation; 3) Usage of a target network. In this section we have ablations to show the contribution of each. The ablations are run with the same setup as the main result, using a batch size of $1408$ unless otherwise specified, but trained only on the train split and evaluated on the validation split.

\begin{figure}[ht]
\begin{center}
\centerline{\includegraphics[width=0.45\columnwidth]{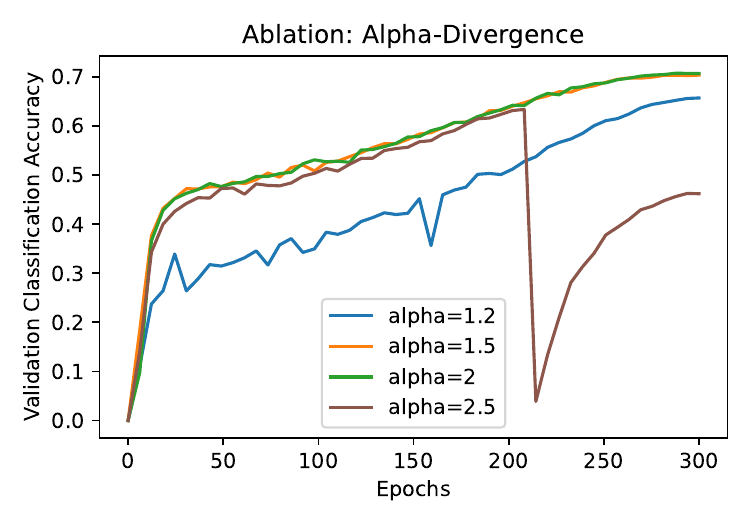}}
\caption{Ablation for the $\alpha$-divergence with auxiliary $\text{EMA}=0$. $\alpha=2$ is the best performing one. $\alpha=1.5$ is close, but smaller $\alpha$ or $\alpha>2$ are much worse.}
\label{fig:alphadivergence}
\end{center}
\vskip -0.2in
\end{figure}

In~\cref{fig:alphadivergence}, we first sweep over various $\alpha$-divergences. For this ablation, the inner scale $s$ is not fixed, but actually learned through a separate SGD+Momentum optimizer with learning rate $0.1$ and momentum $0.9$. Surprisingly, $\alpha=2$ ($\chi^2$-divergence), the simplest Spectral Contrastive version, performs the best. With the larger $\alpha=2.5$, raining becomes noticeably unstable and the performance is dramatically worse. This is because once $\alpha > 2$ there is higher pressure to push some of the embeddings much farther apart from each other, making it easier to diverge. As $\alpha$ gets smaller than $2$ and closer to $1$ ($\alpha=1$ is the KL-divergence version), performance also gets worse. Owing to this result, we fixed $\alpha=2$ (Spectral Contrastive) for our main results and other ablations. This result is surprising at first, due to the results for $f$-MICL~\citep{lu2024f}, which showed much smaller differences between $f$-divergences. On closer inspection, we can see that the objectives of \algabb and $f$-MICL are completely different because we rewrite the contrastive objective using a scalar transformation $t_\alpha$ that changes the second term (the quadratic variance term) into a squared dot product form. This makes \algabb behave very differently than $f$-MICL, and thus perhaps the different $\alpha$-divergences have a much stronger impact on the learned representation.

\begin{figure}[ht]
\begin{center}
\centerline{\includegraphics[width=0.5\columnwidth]{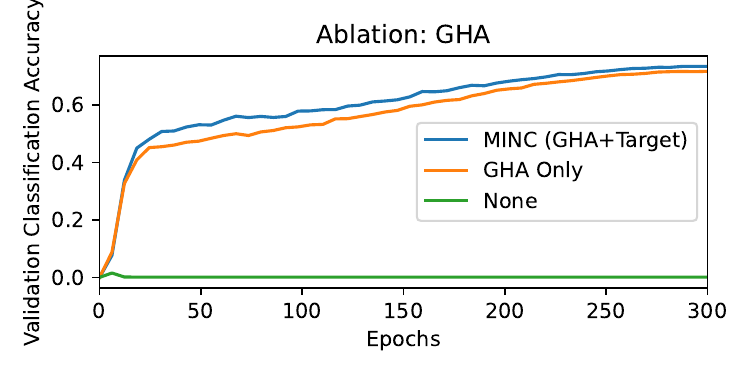}}
\caption{Ablation for GHA when auxiliary $\text{EMA}=0.8$. Without GHA, the representation collapses. With GHA, the representation learns a meaningful representation. Combining GHA and the target network results in even better performance.}
\label{fig:gha}
\end{center}
\vskip -0.2in
\end{figure}

In~\cref{fig:gha}, we investigate the importance of the GHA-derived lower triangular transformation $\texttt{LT}[\cdot]$. We show how just directly using the \algabb objective with an auxiliary EMA of $0.8$ but without a target network nor the GHA transformation completely falls flat; the representation essentially collapses to a constant. Then by adding in the GHA transformation, we see that it no longer collapses and actually learns something. GHA is successfully preventing the collapse that could happen. Finally the complete \algabb with both GHA and target network further improves the performance, showing the further benefit of the target network.

\begin{figure}[ht]
\begin{center}
\centerline{\includegraphics[width=0.85\columnwidth]{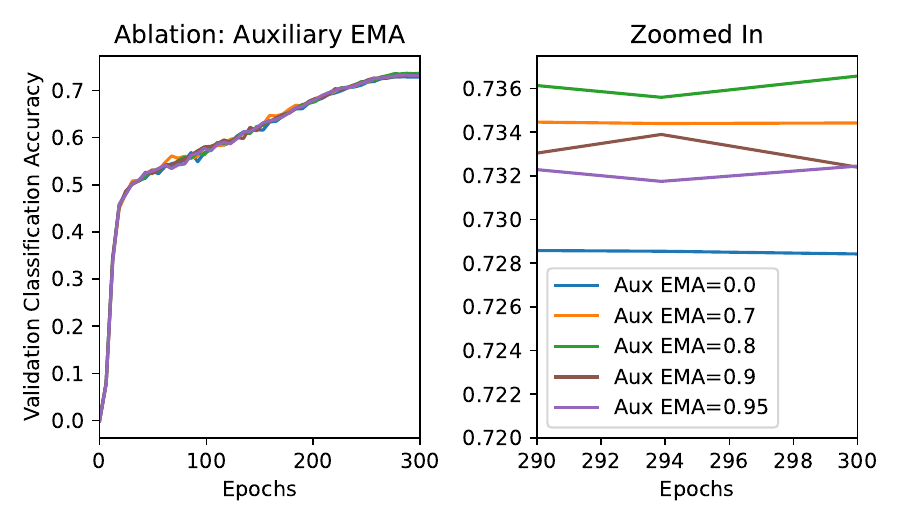}}
\caption{Ablation for auxiliary EMA of \algabb. Any amount of auxiliary EMA is better than 0, with 0.8 being the sweetspot.}
\label{fig:auxema1}
\end{center}
\vskip -0.2in
\end{figure}

The next ablation in~\cref{fig:auxema1} is a sweep over the auxiliary EMA $\beta$. An auxiliary EMA of 0 means that the auxiliary $\Lambda$ is only computed from the current mini-batch, making it very similar to the Spectral Contrastive loss but with GHA and a target network, which performs comparatively the worst, but is still learning a meaningful representation. But increasing the auxiliary EMA further improves the performance, with $0.8$ being the best. Going larger than $0.8$ starts to result in worse performance. The fact that $0.8$ is a sweetspot suggests there is a bias-variance tradeoff due to the non-stationarity of the embeddings $\phi$. A larger EMA means we are accumulating $\Lambda$ over many past mini-batches, which is more biased because the past mini-batches have stale representations. A smaller EMA accumulates over fewer mini-batches, resulting in a less stable but higher variance estimate.

\section{Related Work}
\label{sec:relatedwork}


Our algorithm \algabb starts from the same theoretical framework as the Spectral Contrastive Loss~\citep{haochen2021provable} and $f$-MICL~\citep{lu2024f}, which seek to maximize a notion of mutual information between data points. Other contrastive losses such as InfoNCE~\citep{oord2018representation} and SimCLR~\citep{chen2020simple}, and more~\citep{arora2019theoretical, lee2021predicting, tosh2021contrastive} do not fall into the exact same framework, but are still similar in that the goal is still to maximize some notion of mutual information. Because contrastive methods need to compare every data point to every other data point, their objective have a quadratic dependence on the size of the dataset, which results in high variance. Our work builds on this mutual information framework, and goes on to derive a non-contrastive version based on the theoretical ideas of power iteration in order to reduce this variance dependence.


On the non-contrastive side, BYOL~\citep{grill2020bootstrap} amongst many others~\citep{chen2021exploring,bardes2021vicreg,zbontar2021barlow,liu2022bridging, tian2021understanding, wang2021towards} have been exploring ways to remedy the quadratic dependence on the dataset. Our \algabb algorithm, is a non-contrastive version of the Spectral Contrastive loss that does not have the quadratic dependence. Interestingly, through a power iteration viewpoint, \algabb can in fact be related to a linear predictor version of BYOL (\cref{sec:moretheory}). This connection brings non-contrastive methods and contrastive methods closer together, in that they all can be seen to be ultimately related to maximizing mutual information, just in somewhat different ways.



\section{Conclusion}

Our~\algabb algorithm successfully improves upon the Spectral Contrastive loss by turning it from a contrastive loss into a non-contrastive loss. Theoretically, we provide a power iteration perspective that can relate contrastive and non-contrastive losses, connecting the Spectral Contrastive loss and the Linear BYOL method together. In ImageNet, \algabb consistently shows an improvement over the Spectral Contrastive loss, and reduces the gap between contrastive methods and BYOL, the (non-contrastive) state-of-the-art. Interestingly, we found that~\algabb performs on par with Linear BYOL, which suggests that the non-linear predictor in BYOL can inspire further improvements to contrastive and non-contrastive spectral losses.
More broadly, beyond ImageNet, our work does show the feasibility of combining the strengths of contrastive and non-contrastive objectives together, and could pave the way for more powerful contrastive to non-contrastive conversions in the future.
On the theory side, the power iteration perspective may also prove a useful tool to strengthen existing results and weaken some of the limiting assumptions.

\bibliographystyle{abbrvnat}
\nobibliography*
\bibliography{minc}


%







\end{document}